\documentclass{article}

    \usepackage[preprint]{neurips_2024}

\usepackage[utf8]{inputenc} % allow utf-8 input
\usepackage[T1]{fontenc}    % use 8-bit T1 fonts
\usepackage{hyperref}       % hyperlinks
\usepackage{url}            % simple URL typesetting
\usepackage{booktabs}       % professional-quality tables
\usepackage{amsfonts}       % blackboard math symbols
\usepackage{nicefrac}       % compact symbols for 1/2, etc.
\usepackage{microtype}      % microtypography
\usepackage{xcolor}         % colors

\usepackage{amsmath}
\usepackage{amsthm}
\usepackage[ruled,vlined,linesnumbered,algo2e]{algorithm2e}
\usepackage{algorithm}
\usepackage{graphicx}

\usepackage{amssymb}

\usepackage{tabularx}
\usepackage[table]{xcolor}
\usepackage{xcolor}
\usepackage{soul}
\usepackage{subcaption}
\usepackage{float}

\title{AC-SINDy: Compositional Sparse Identification of Nonlinear Dynamics}

\author{%
  Peter Racioppo \\
  Independent Researcher \\
  Los Angeles, CA 90731 \\
  \texttt{pcracioppo@gmail.com} \\
}

\begin{document}

\maketitle

\begin{abstract}
We present AC-SINDy, a compositional extension of the Sparse Identification of Nonlinear Dynamics (SINDy) framework that replaces explicit feature libraries with a structured representation based on arithmetic circuits. Rather than enumerating candidate basis functions, the proposed approach constructs nonlinear features through compositions of linear functions and multiplicative interactions, yielding a compact and scalable parameterization and enabling sparsity to be enforced directly over the computational graph. We also introduce a formulation that separates state estimation from dynamics identification by combining latent state inference with shared dynamics and multi-step supervision, improving robustness to noise while preserving interpretability. Experiments on nonlinear and chaotic systems demonstrate that the method recovers accurate and interpretable governing equations while scaling more favorably than standard SINDy.
\end{abstract}

\section{Introduction}

Discovering governing equations of dynamical systems from data is a central problem in scientific machine learning. The Sparse Identification of Nonlinear Dynamics (SINDy) framework addresses this by selecting a sparse combination of candidate basis functions from a predefined library \citep{Brunton_2016}, enabling recovery of compact and interpretable models.

However, SINDy suffers from a mismatch between representation and structure. Many dynamical systems exhibit sparse, low-order interactions, yet SINDy represents them by enumerating a large library of candidate functions. This library grows combinatorially with state dimension and interaction order, even when only a small number of terms are active.

We replace sparse selection in a fixed basis with sparse structure in a learned computational graph. Rather than enumerating all monomials up to a given degree, the proposed approach constructs features through compositions of linear functions and multiplicative interactions, yielding a compact representation in which individual interactions can be expressed with a number of parameters that scales linearly in the state dimension.

This induces a shift in the underlying hypothesis class: instead of assuming sparsity in a predefined basis, we assume that the dynamics admit a low-complexity compositional representation. Concretely, AC-SINDy approximates functions that admit low-rank polynomial tensor structure (via factorization) or, equivalently, low-complexity arithmetic circuits, rather than sparse monomial expansions. As a result, functions that require many monomials in an explicit expansion may still admit compact representations under this compositional parameterization.

Although implemented using a neural architecture, the model is more appropriately understood as a structured computational graph (arithmetic circuit) whose parameters correspond to edges. Pruning removes identifiable components of this graph, yielding a sparse structure that can be interpreted symbolically.

We also address the challenge of learning dynamics from noisy observations. Rather than estimating derivatives or explicitly denoising the signal, we treat state estimation and dynamics learning as a coupled problem. We introduce a learned filtering stage that produces a latent state, which is then evolved by an interpretable dynamics model trained with multi-step supervision. The key observation is that accurate prediction requires a denoised representation of the state. By enforcing consistency over multiple time steps, the model is encouraged to learn representations that are maximally predictive under the dynamics, which implicitly suppresses noise. This yields a separation between flexible state estimation and interpretable dynamics identification, while enabling end-to-end training.

We further introduce \emph{Feature Normalization}, a scale-invariant parameterization that ensures learned coefficients reflect functional importance rather than feature magnitude. This is particularly important in compositional models with multiplicative interactions, where feature scales can be arbitrarily redistributed across layers without changing the represented function, leading to non-identifiability of coefficients.

Finally, we enforce sparsity through iterative pruning using gradient-based importance estimates, enabling structure learning in the shared computational graph.

We evaluate the method on a range of nonlinear and chaotic dynamical systems, including settings with noisy observations. Results demonstrate that the approach recovers accurate and interpretable models while scaling more favorably than standard SINDy.

Our contributions are as follows:
\begin{itemize}
    \item We introduce AC-SINDy, which replaces sparse basis selection with sparse structure in a learned computational graph.

    \item We propose a formulation that separates filtering from dynamics identification via latent state estimation and multi-step consistency, enabling robust learning from noisy data.

    \item We introduce Feature Normalization, a scale-invariant parameterization that stabilizes optimization while retaining interpretability in compositional models.

    \item We develop a pruning-based structure learning procedure for shared computational graphs and demonstrate recovery of sparse, interpretable dynamics across nonlinear and chaotic systems.
\end{itemize}

This work represents an initial investigation of compositional, factorized representations for system identification; future work will explore scaling to higher-dimensional systems and more complex dynamics.

\section{Related Work}

The discovery of governing equations from data has been significantly advanced by the Sparse Identification of Nonlinear Dynamics (SINDy) framework \citep{Brunton_2016}, which employs sparse regression over a fixed library of candidate functions to obtain interpretable models. However, this approach is sensitive to noise, depends on the choice of coordinates, and suffers from combinatorial growth in library size.

Several extensions have addressed these limitations. \citet{Kaheman_2020} introduced SINDy-PI to handle implicit and rational dynamics. \citet{Champion_2019} proposed the SINDy Autoencoder, which learns a latent coordinate system in which the dynamics are sparse, demonstrating that neural networks can aid in discovering parsimonious representations.

More recent work has begun to relax the fixed-library assumption by learning or adapting candidate functions directly from data. These approaches include parameterized or learned libraries \citep{Yonezawa_2026, singh2024learnlearnableadaptiverepresentations}, as well as architectural extensions such as nested or hierarchical SINDy formulations \citep{Fiorini_2025, STEIGER2025427}. In parallel, advances in neural symbolic regression and compositional architectures, such as Kolmogorov–Arnold Networks, demonstrate the potential of structured, compositional function representations \citep{faroughi2025scientificmachinelearningkolmogorovarnold}. These directions are closely related to tensor factorization approaches, where nonlinear interactions are represented via low-rank decompositions of coefficient tensors, providing a compact alternative to explicit basis expansion \citep{tensor_decomp}.

Despite these advances, most existing approaches retain an explicit or parameterized function library, and continue to perform sparse regression within this basis.

In contrast, our approach replaces explicit libraries with a compositional representation in which features are constructed through learned compositions of linear transformations and multiplicative interactions. This defines an implicit hypothesis class and enforces sparsity over the structure of the resulting computational graph, rather than over coefficients in a predefined basis.

From a functional perspective, this formulation is closely related to low-rank tensor methods and arithmetic circuit representations, which provide compact representations of high-dimensional functions through factorization and composition. In particular, the model can be interpreted as learning a factorized representation of polynomial interactions, analogous to a low-rank (CP) decomposition of the underlying coefficient tensor. Unlike prior SINDy-based approaches, which operate in an explicit or learned feature basis, AC-SINDy operates directly in this compositional regime, enabling improved parameter efficiency and scalability.

A central challenge in SINDy is estimating time derivatives from noisy data. Early work improved robustness through regularization and numerical differentiation \citep{rudy2016datadrivendiscoverypartialdifferential}. Subsequent approaches have replaced explicit differentiation with learned surrogates: \citet{Both_2021} introduced DeePyMoD, which estimates derivatives via automatic differentiation within neural networks, while \citet{forootani2023robustsindyapproachcombining} replace differential formulations with integral constraints to reduce sensitivity to noise. Related work has also combined SINDy with Neural Ordinary Differential Equations (Neural ODEs), enabling multi-step training and temporal consistency through backpropagation in time \citep{lee2021structurepreservingsparseidentificationnonlinear}.

These approaches focus on stabilizing derivative estimation or explicitly denoising the signal prior to system identification. In contrast, we learn a latent state representation jointly with the dynamics model using multi-step predictive consistency. This encourages representations that are robust to noise while preserving interpretability in the learned dynamics.

% Recent work has also explored incorporating compositional structure directly into the SINDy framework through deep or nested architectures. These approaches construct hierarchical models by stacking SINDy layers or combining library-based transformations with neural network components, enabling representation of more complex functional forms. However, they retain an explicit or parameterized dictionary of candidate functions, and continue to rely on sparse regression within this basis. In contrast, our approach eliminates the notion of a predefined function library entirely, replacing it with a learned compositional representation that operates directly on the state variables.

% You can strengthen your positioning by lightly referencing:
% Sum-Product Networks (SPNs)
% → also arithmetic circuits with probabilistic semantics
% Tensor factorizations
% → your bilinear structure is essentially CP decomposition
% Kolmogorov–Arnold representations
% → compositional function structure

\section{Background: The SINDy Algorithm}

Consider a first-order nonlinear dynamical system of the form
\[
\dot{x}(t) = f(x(t), u(t), t),
\]
where $x(t) \in \mathbb{R}^n$ is the state vector and $u(t)$ denotes control inputs. The Sparse Identification of Nonlinear Dynamics (SINDy) framework seeks to approximate the unknown function $f$ using a sparse linear combination of candidate basis functions.

\begin{figure}[H]
\begin{center}
\includegraphics[width=0.8 \textwidth]{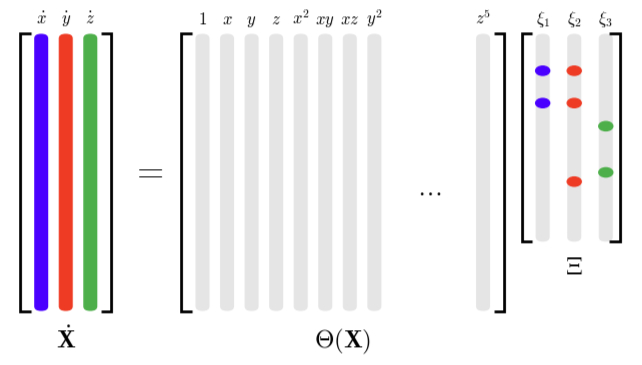}
\caption{Illustration of SINDy regression showing sparse identification over a candidate library, adapted from \citet{Brunton_2016}.}
\label{fig:SINDy}
\end{center}
\end{figure}

Specifically, for each component $x_i(t)$, $i \in \{1, \dots, n\}$, SINDy solves a regression problem of the form
\[
\mathcal{L}_i = \left\| \dot{x}_i(t) - g_i(x(t), t) \right\|_2^2,
\]
where
\[
g_i(x(t), t) = \sum_{j} w_{i,j} f_{i,j}(x(t), t)
\]
is a linear combination of predefined basis functions $\{f_{i,j}\}$. The key assumption is that only a small subset of these basis functions is required to accurately describe the system dynamics. The SINDy algorithm is illustrated in Fig.~\ref{fig:SINDy}.

In practice, SINDy performs least-squares regression combined with iterative thresholding, sequentially removing coefficients with small magnitude (typically based on an L1 criterion). Empirically, this procedure has been observed to outperform standard LASSO-based approaches in identifying sparse and interpretable models. 

\section{AC-SINDy: Compositional Sparse Identification via Arithmetic Circuits}

We propose a compositional extension of the SINDy framework in which the explicit library of candidate functions is replaced by a structured neural architecture. The model constructs nonlinear features through compositions of learned transformations, yielding a compact and scalable representation of dynamical systems.

This model can be interpreted as learning a restricted, factorized arithmetic circuit representation of the dynamics. Rather than enumerating a basis of candidate functions, the model represents the target function as a directed acyclic graph of additions and multiplications, enabling reuse of intermediate computations. 

This perspective highlights a key distinction from standard SINDy. While SINDy assumes sparsity in a fixed basis of monomials, the proposed approach assumes that the dynamics admit a low-complexity compositional (circuit) representation. A function may require many monomials in an explicit expansion while still admitting a compact factorized representation. The proposed formulation therefore replaces sparsity in a predefined basis with low compositional complexity as the primary inductive bias.

While the architecture reduces to an arithmetic circuit when restricted to linear and multiplicative operations, it extends naturally to more general compositional models by incorporating additional nonlinear primitives.

\subsection{Architecture}

Let $x \in \mathbb{R}^n$ denote the system state. At each layer $\ell$, we form an augmented input
\[
\tilde{x}^{(\ell)} = [x^{(\ell)}; 1] \in \mathbb{R}^{n_\ell + 1},
\]
where the constant feature enables representation of bias and lower-order terms.

Each layer consists of:

\paragraph{Masked Linear layer.}
\[
h^{(\ell)} = W^{(\ell)} \tilde{x}^{(\ell)}, \quad W^{(\ell)} \in \mathbb{R}^{d_\ell \times (n_\ell + 1)}.
\]

\paragraph{Interaction layer.}
\[
x^{(\ell+1)}_k = \prod_{j \in \mathcal{G}_k} h^{(\ell)}_j,
\]
where $\{\mathcal{G}_k\}$ partitions the features. More general nonlinear primitives may also be used.

Stacking layers yields a compositional computation graph in which features are constructed recursively, and symbolic expressions can be recovered by propagating algebraic structure through the network. Each parameter corresponds to an edge in this computational graph, and learned structure emerges through pruning. This distinguishes the model from standard neural networks, where parameters do not map cleanly to identifiable functional components.

\subsection{Function Class}

The model represents functions of the form:
\[
f(x) = \sum_{k=1}^m \psi_k(x), \qquad
\psi_k(x) = \prod_{j \in S_k} (w_{k,j}^T x).
\]

This corresponds to a sum of factored terms, enabling shared structure across interactions. In contrast to explicit polynomial expansions, where each monomial is represented independently, this representation provides a compact factorization of higher-order interactions.

Unlike standard SINDy, which uses a fixed library $\Theta(x)$, the model defines an implicit hypothesis class through composition. From this perspective, the model can be interpreted as learning a low-rank factorization of the coefficient tensor of a multivariate polynomial, analogous to a CP decomposition. This provides a compact representation of higher-order interactions without explicit enumeration.

\subsection{Expressivity and Sparse Compositional Structure}

The architecture does not span all polynomial terms simultaneously. For example, a quadratic function
\[
f(x) = x^T Q x
\]
is represented as
\[
f(x) = \sum_{k=1}^r (a_k^T x)(b_k^T x),
\]
corresponding to a low-rank factorization of $Q$.

More generally, the model replaces explicit basis expansion with factorized representations. Functions that require $\mathcal{O}(n^p)$ monomials in an explicit expansion may admit representations using $\mathcal{O}(n p)$ parameters when expressed compositionally. This reflects a known separation: some functions require exponentially many monomials in an explicit expansion but admit polynomial-size representations as arithmetic circuits. In this sense, AC-SINDy targets a different hypothesis class: functions that admit low-rank tensor structure or low arithmetic circuit complexity, rather than sparsity in a fixed monomial basis.

Accordingly, the model assumes that the underlying dynamics admit a low-complexity compositional structure rather than sparsity in a fixed basis.

\subsection{Interpretation as an Arithmetic Circuit}

When restricted to linear and multiplicative operations, the model defines a restricted arithmetic circuit with a sum-of-products (factorized) structure:
\begin{itemize}
    \item Linear layers correspond to sum nodes
    \item Multiplicative layers correspond to product nodes
    \item Pruning removes edges in the computational graph
\end{itemize}

This enables reuse of intermediate features and yields a compact, factorized representation of the dynamics.

\subsection{Sparsification}

Sparsity is enforced by pruning weights in the linear layers. Each parameter corresponds to an edge in the computational graph, and pruning removes edges with minimal impact on performance.

Unlike standard SINDy, sparsity operates over both feature construction and composition. As a result, sparsity corresponds to structural simplicity of the learned computational graph, rather than selection of terms from a predefined basis.

\subsection{Parameter Efficiency}

In standard SINDy, polynomial libraries grow combinatorially with the state dimension $d$. The number of monomials of degree $p$ in $d$ state variables is:
\[
\binom{d + p - 1}{p}, \quad \text{and} \quad
\sum_{k=0}^{p} \binom{d + k - 1}{k} \sim \mathcal{O}(p d^p).
\]

In contrast, each compositional term requires $\mathcal{O}(p d)$ parameters, yielding $\mathcal{O}(m p d)$ total parameters, where $m$ is the number of active compositional terms after pruning.

AC-SINDy provides improved scaling when the number of active compositional terms satisfies $m < d^{p-1}$, i.e., when the number of interactions grows more slowly than the combinatorial number of $(p-1)$-order terms.

\begin{figure}[H]
     \centering
     \begin{subfigure}[b]{0.32\textwidth}
         \centering
         \includegraphics[width=\textwidth]{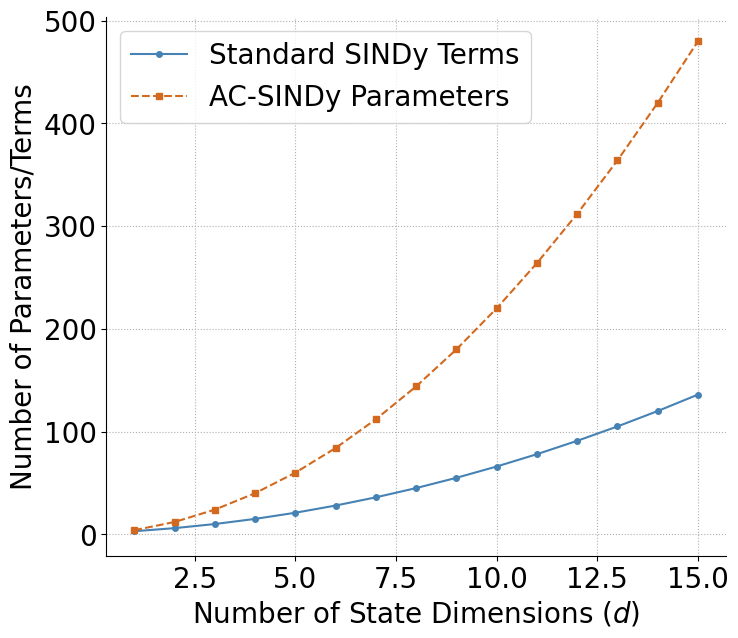}
         \caption{$p=2$}
         \label{fig:sindy_p2}
     \end{subfigure}
     \hfill
     \begin{subfigure}[b]{0.32\textwidth}
         \centering
         \includegraphics[width=\textwidth]{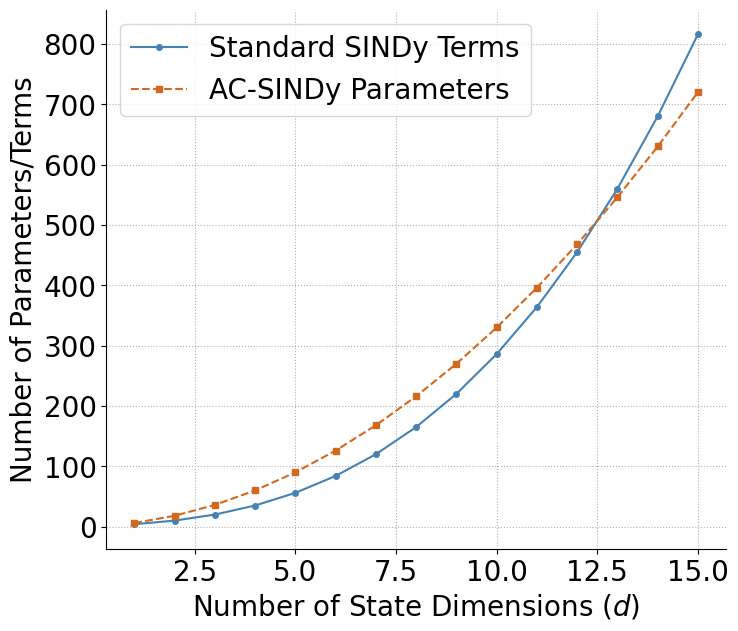}
         \caption{$p=3$}
         \label{fig:sindy_p3}
     \end{subfigure}
     \hfill
     \begin{subfigure}[b]{0.32\textwidth}
         \centering
         \includegraphics[width=\textwidth]{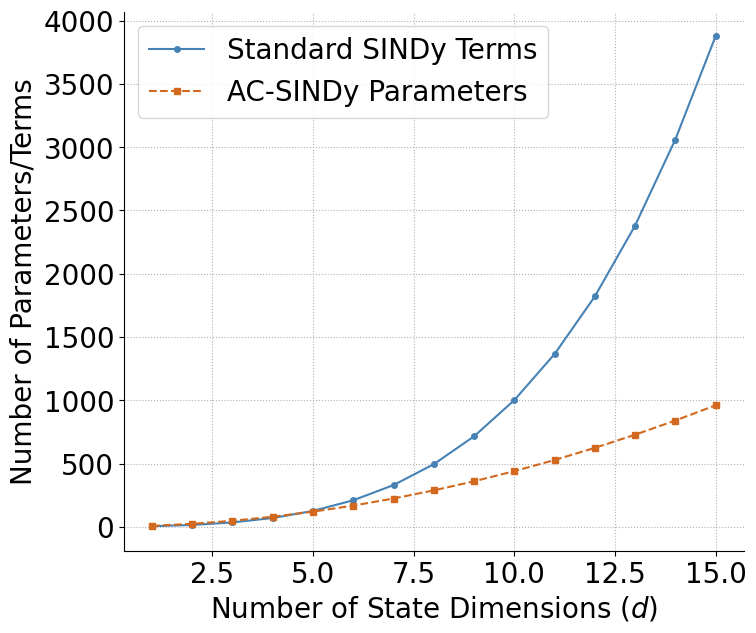}
         \caption{$p=4$}
         \label{fig:sindy_p4}
     \end{subfigure}
        \caption{Parameter growth for standard SINDy (explicit enumeration) versus AC-SINDy (compositional representation) for polynomial orders $p=2,3,4$.}
        \label{fig:sindy_vs_neural}
\end{figure}

In many physical systems, interactions scale linearly with the state dimension ($m = \mathcal{O}(d)$), yielding $\mathcal{O}(p d^2)$ growth. Figure~\ref{fig:sindy_vs_neural} illustrates this regime. For $p=2$, both methods scale as $\mathcal{O}(d^2)$ (with SINDy having a smaller constant), while for higher orders SINDy grows combinatorially and AC-SINDy remains quadratic.

\section{Importance-Based Pruning in Compositional Models}

SINDy enforces sparsity by selecting a small number of active terms from a predefined library. In this setting, each parameter corresponds to a single basis function, and coefficient magnitude provides a reasonable proxy for importance. 

In AC-SINDy, however, parameters contribute jointly through shared intermediate computations, and importance must be understood in terms of their effect on the overall computation graph rather than as standalone coefficients. Parameters correspond to edges that may contribute to many compositional terms, breaking the one-to-one correspondence between parameters and functions. As a result, magnitude-based sparsification is no longer well-aligned with functional importance.

To address this, we consider the marginal contribution of each parameter,
\[
\Delta_i = \mathcal{L}(w_{-i}) - \mathcal{L}(w),
\]
which measures the effect of removing a parameter on the loss. While this provides a principled notion of importance, computing it exactly is not scalable.

Instead, we approximate this quantity using a first-order Taylor expansion,
\[
\Delta_i \approx \left| w_i \frac{\partial \mathcal{L}}{\partial w_i} \right|,
\]
yielding a gradient-based pruning criterion. This approximation captures both the magnitude of a parameter and its influence on the loss, while remaining computationally efficient.

In practice, we track a moving average of the training loss and trigger pruning once optimization plateaus. At each pruning step, a subset of parameters with the lowest estimated importance is removed, after which the model is finetuned to recover performance. Repeating this process produces a sequence of progressively sparser models, from which the final model is selected based on validation loss.

% https://polukhin.tech/2022/10/27/pruning-the-history-and-overview?utm_source=chatgpt.com#pruning-methods
\section{Feature Normalization}

A key challenge in both SINDy and compositional models is sensitivity to feature scaling. Candidate functions such as $x$, $x^2$, and higher-order terms can differ significantly in magnitude depending on the region of the state space. As a result, learned coefficients depend inversely on feature scale, making coefficient magnitude an unreliable indicator of functional importance and distorting sparsity-based model selection.

This issue is exacerbated in compositional models with multiplicative interactions. Feature scales can be arbitrarily redistributed across layers without changing the output, leading to non-identifiability: multiple parameter configurations represent the same function while assigning different magnitudes to coefficients.

To address this, we introduce \emph{Feature Normalization}, which rescales each feature independently using data-dependent statistics. Let $h(x) \in \mathbb{R}^d$ denote the feature vector produced by the model. We define
\[
\tilde{h}_i(x) = \frac{h_i(x)}{\sigma_i + \epsilon},
\]
where $\sigma_i$ is the standard deviation of feature $h_i$ estimated over the training trajectory. The model output is then
\[
f(x) = W \tilde{h}(x).
\]

In contrast to standard normalization methods, Feature Normalization preserves identifiability. Methods such as Batch Normalization and Layer Normalization introduce learnable affine transformations that allow features to be rescaled without changing the represented function. In compositional models, this induces a large class of equivalent parameterizations, breaking the correspondence between coefficient magnitude and functional importance.

Feature Normalization removes this degree of freedom by fixing feature scales using statistics treated as constants (via stop-gradients). As a result, it yields a scale-invariant parameterization in which coefficient magnitudes reflect functional contribution.

In practice, we compute $\sigma_i$ using running statistics and detach it from the computational graph:
\[
\tilde{h}_i(x) = \frac{h_i(x)}{\text{stopgrad}(\sigma_i) + \epsilon}.
\]

\section{Multi-Step Consistency and Separation of Filtering and Dynamics}
\label{sec:multi_step_filter}

A central challenge in learning dynamical systems from data is disentangling state estimation from dynamics identification. In many settings, observations are noisy or partial, and models trained on one-step prediction can implicitly combine filtering and dynamics into a single function. While such models may fit the data, they are not constrained to represent a time-consistent dynamical system.

To address this, we adopt a formulation that enforces dynamical consistency by requiring a single learned function to be closed under repeated composition over time. Let $S_t$ denote a latent state estimate and $f_\theta$ a learned dynamics model. We define the evolution
\[
S_{t+1} = S_t + f_\theta(S_t)\,\Delta t, \qquad
S_{t+k} = \Phi_\theta^{(k)}(S_t),
\]
where $\Phi_\theta^{(k)}$ denotes $k$ successive applications of $f_\theta$. This shared-parameter structure enforces consistency under composition.

Crucially, rollout is combined with multi-step supervision,
\[
\mathcal{L} = \sum_{k=1}^{K} \left\| \hat{x}_{t+k} - x_{t+k} \right\|^2,
\]
which enforces accuracy over extended horizons. Without this constraint, an expressive encoder can encode future information into $S_t$, allowing trivial dynamics to minimize one-step loss. Multi-step supervision eliminates this degeneracy by requiring consistent predictions across multiple time steps.

Under this formulation, the model decomposes into a filtering component and a dynamics component:
\[
S_t = \mathcal{E}(x_{1:t}), \qquad
S_{t+k} = \Phi_\theta^{(k)}(S_t).
\]
The encoder $\mathcal{E}$ estimates the latent state from observations, while $f_\theta$ governs its evolution. Because $f_\theta$ is applied repeatedly and evaluated over multiple steps, it must represent a self-consistent dynamical law rather than relying on access to the observation history.

This structure aligns with classical system identification, where a state estimate is propagated using a fixed model. In contrast to one-step regression, the multi-step formulation enforces temporal consistency and discourages solutions that do not generalize under rollout.

% \section{Control with Structured Dynamics}

% The proposed framework naturally supports model-based control by providing an explicit, interpretable representation of the system dynamics. Given a learned model
% \[
% \dot{x} = g_w(x, u),
% \]
% standard control techniques can be applied, including linearization and trajectory optimization.

% A key advantage of AC-SINDy over black-box neural models is the structured and sparse form of $g_w$. This enables direct reasoning about the dynamics and, in some cases, closed-form controller design. For example, consider a system of the form
% \[
% \dot{x} = p_1(x) + k u,
% \]
% where $p_1(x)$ is approximated by a sparse polynomial. If the desired dynamics are $\dot{x} = p_2(x)$, and the control input is parameterized as a polynomial $u = q_\theta(x)$, then the closed-loop system remains polynomial, and the controller parameters can be chosen to match coefficients:
% \[
% \theta_i \approx \frac{b_i - a_i}{k}.
% \]

% More generally, the compositional structure of the learned model allows for efficient local linearization and application of classical control methods such as LQR. Compared to dense neural representations, the sparse and structured form of AC-SINDy models may lead to more stable and interpretable control policies.

% We leave a detailed empirical investigation of control applications to future work.

\section{Experimental Results}

We implemented the proposed AC-SINDy architecture in PyTorch, including compositional feature construction, gradient-based pruning, and multi-step training for dynamical consistency.\footnote{\url{https://github.com/PCR-git/AC-SINDy}}

\subsection{Experimental Setup}

We evaluate the method on low-dimensional dynamical systems, including canonical chaotic and nonlinear systems.

\paragraph{2D nonlinear system.}
\[
\dot{x}(t) = -0.1x + y, \quad
\dot{y}(t) = -2x - 0.1y - 0.5xy - 0.025y^2.
\]

\paragraph{Lorenz system.}
\[
\dot{x} = \sigma (y - x), \quad
\dot{y} = x (\rho - z) - y, \quad
\dot{z} = x y - \beta z,
\]
with $(\sigma, \rho, \beta) = (10, 28, 8/3)$.

\paragraph{Lorenz system with sinusoidal forcing.}
\[
\dot{x}(t) = \sigma (y - x) + 0.1 \sin(x), \quad
\dot{y}(t) = x (\rho - z) - y, \quad
\dot{z}(t) = x y - \beta z.
\]

\subsection{Training Dynamics}

\begin{figure}[ht]
\centering
\includegraphics[width=0.6\textwidth]{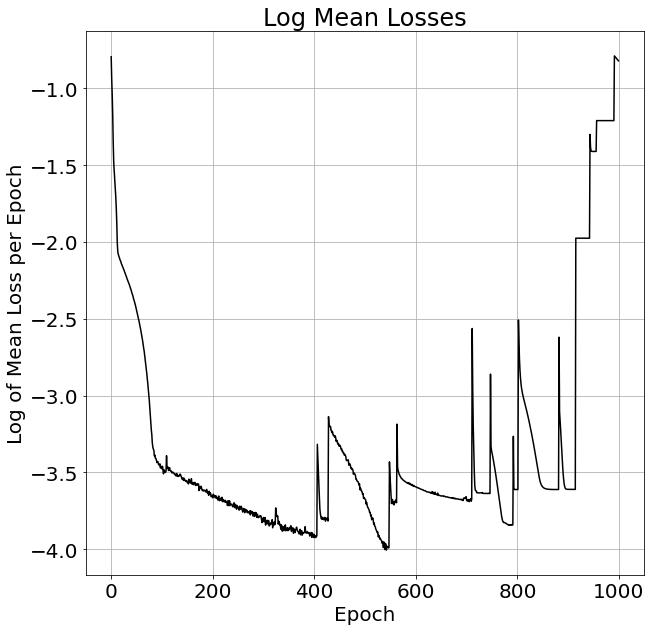}
\caption{Training loss during iterative pruning. Loss spikes occur when parameters are removed, followed by recovery as the model re-optimizes. Excessive pruning degrades performance.}
\label{fig:loss}
\end{figure}

Figure~\ref{fig:loss} shows the training loss during iterative pruning. Each pruning step induces a temporary increase in loss, followed by recovery during retraining. For systems with sparse underlying dynamics, pruning initially improves performance by removing redundant parameters, but eventually leads to underfitting. The final model is selected at the minimum validation loss, corresponding to the optimal sparsity--expressivity tradeoff.

\subsection{Prediction Quality}

\begin{figure}[ht]
\centering
\includegraphics[width=0.48\textwidth]{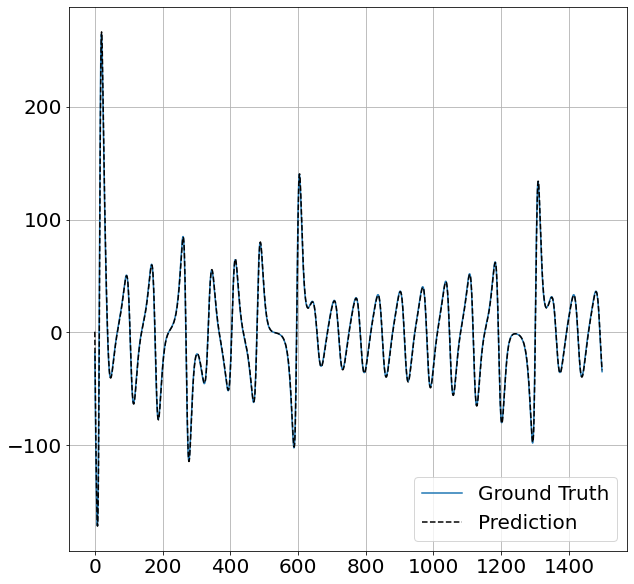}
\includegraphics[width=0.48\textwidth]{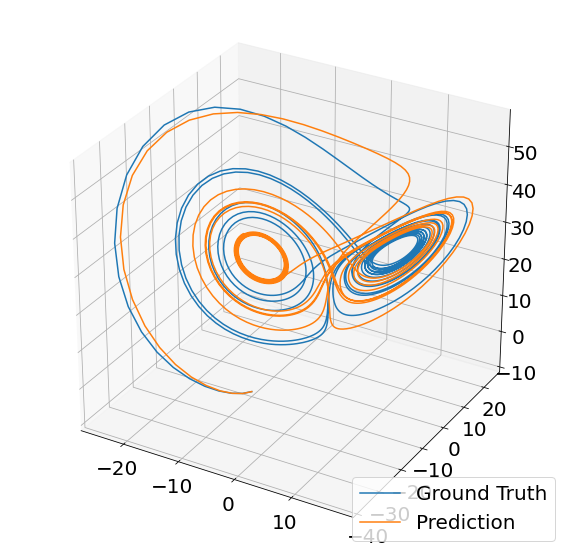}
\caption{Left: Predicted vs. true trajectory for a single state variable. Right: Phase-space trajectory. The learned dynamics capture the qualitative structure despite long-term divergence.}
\label{fig:predictions}
\end{figure}

Figure~\ref{fig:predictions} shows predictions on the Lorenz system. The model accurately captures both individual state dynamics and the global phase-space structure. As expected for chaotic systems, long-term trajectories diverge from the ground truth.

\subsection{Model Recovery}

A primary objective of SINDy-based methods is recovery of the underlying governing equations. The proposed approach identifies sparse and interpretable structure, including nonlinear interaction terms.

\paragraph{2D nonlinear system.}
\[
\text{True dynamics:} \quad \dot{x} = -0.1x + y, \quad
\dot{y} = -2x - 0.1y - 0.5xy - 0.025y^2.
\]
\[
\text{Recovered system:} \quad \dot{x} \approx -0.11x + 1.00y, \quad
\dot{y} \approx -1.93x - 0.11y - 0.50xy - 0.029y^2.
\]

\paragraph{Lorenz system.}
\[
\text{True dynamics:} \quad \dot{x} = 10(y - x), \quad
\dot{y} = x(28 - z) - y, \quad
\dot{z} = xy - \tfrac{8}{3}z.
\]
\[
\text{Recovered system:} \quad
\begin{aligned}
\dot{x} &\approx 8.72\,y - 7.17\,x, \quad
\dot{y} \approx 26.18\,x + 0.90\,y, \\
\dot{z} &\approx 0.25\,x^2 + 0.49\,xy + 0.10\,x + 0.25\,y^2 - 0.03\,y - 7.18.
\end{aligned}
\]

The learned 2D nonlinear model recovers the correct functional structure. For the Lorenz system, while the exact coefficients are not perfectly recovered and some spurious terms appear, the dominant interactions and qualitative dynamics are correctly identified.

These results demonstrate that the proposed approach can recover accurate dynamical models despite not explicitly enumerating candidate terms, demonstrating that compositional parameterization is sufficient for sparse system identification.

\subsection{Noisy Data and Filtering}

We evaluate the proposed approach in a noisy setting using the filtering-based formulation described in Section~\ref{sec:multi_step_filter}. We consider the same 2D nonlinear system:
\[
\dot{x} = -0.1x + y, \quad
\dot{y} = -2x - 0.1y - 0.5xy - 0.025y^2,
\]
and add Gaussian noise to the observed trajectories:
\[
\tilde{x}(t) = x(t) + \epsilon(t), \quad \epsilon \sim \mathcal{N}(0, 0.05^2).
\]

\begin{figure}[ht]
\centering
\includegraphics[width=0.7\textwidth]{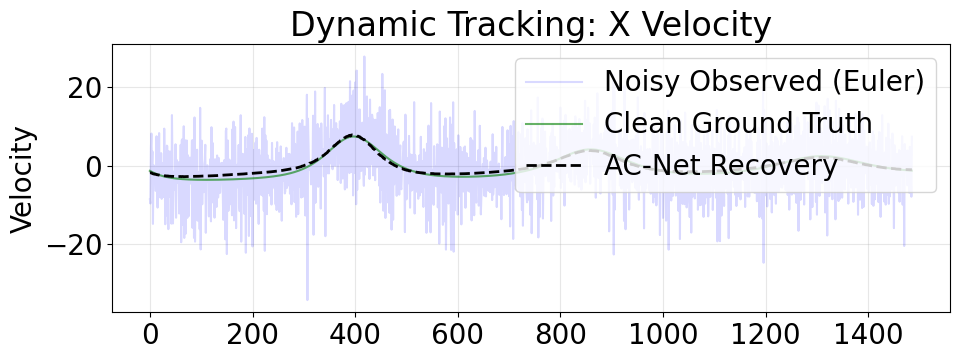}
\\
\includegraphics[width=0.7\textwidth]{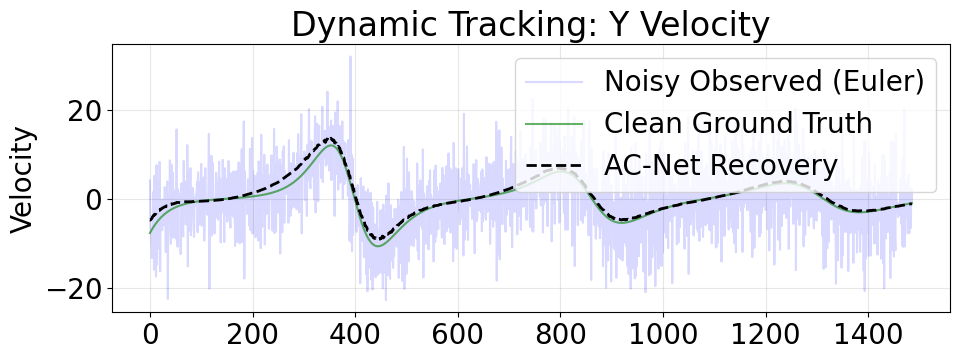}
\caption{Noisy system identification with Gaussian noise ($\sigma = 0.05$). The model fits the noisy observations while recovering a smooth approximation of the underlying dynamics.}
\label{fig:noisy}
\end{figure}

Figure~\ref{fig:noisy} shows the noisy observations, ground truth trajectories, and model predictions. Despite significant observation noise, the model is able to fit the trajectories and recover a smooth approximation of the underlying dynamics.

The recovered governing equations are:
\[
\text{Recovered system:} \quad
\dot{x} \approx -0.08\,x + 0.55\,y,\quad
\dot{y} \approx 0.04\,x^2 - 0.57\,xy - 2.14\,x.
\]

The recovered model closely matches the dominant linear and bilinear terms, while introducing a small spurious quadratic term ($x^2$), omitting the weaker $y^2$ interaction, and exhibiting minor coefficient deviations. 

\section{Conclusion}

We introduced AC-SINDy, a compositional reformulation of SINDy that replaces explicit feature libraries with a structured, factorized representation of nonlinear dynamics. By constructing features through compositions of simple primitives and enforcing sparsity over the resulting computational graph, the method enables interpretable model discovery without explicit basis enumeration. Experiments demonstrate that the approach can recover accurate and compact governing equations.

% \section{Additional Ideas}

% 1. Loss-mediated modularity
% Components specialize via how they interact with the loss
% Very underexplored.

% 2. AC-SINDy = sparse low-rank operator learning
% Not feature selection, but factorized operator discovery
% This is your best reframing.

% 3. Shared structure between inference (RFA) and dynamics
% One model governs both filtering and evolution
% This is very novel.

% 4. Sparse performs better than dense even on the train set.
% Optimization + factorization bias
% This is worth highlighting experimentally.

\bibliographystyle{plainnat}
\bibliography{references}

\begin{thebibliography}{13}
\providecommand{\natexlab}[1]{#1}
\providecommand{\url}[1]{\texttt{#1}}
\expandafter\ifx\csname urlstyle\endcsname\relax
  \providecommand{\doi}[1]{doi: #1}\else
  \providecommand{\doi}{doi: \begingroup \urlstyle{rm}\Url}\fi

\bibitem[Both et~al.(2021)Both, Choudhury, Sens, and Kusters]{Both_2021}
Gert-Jan Both, Subham Choudhury, Pierre Sens, and Remy Kusters.
\newblock Deepmod: Deep learning for model discovery in noisy data.
\newblock \emph{Journal of Computational Physics}, 428:\penalty0 109985, March 2021.
\newblock ISSN 0021-9991.
\newblock \doi{10.1016/j.jcp.2020.109985}.
\newblock URL \url{http://dx.doi.org/10.1016/j.jcp.2020.109985}.

\bibitem[Brunton et~al.(2016)Brunton, Proctor, and Kutz]{Brunton_2016}
Steven~L. Brunton, Joshua~L. Proctor, and J.~Nathan Kutz.
\newblock Discovering governing equations from data by sparse identification of nonlinear dynamical systems.
\newblock \emph{Proceedings of the National Academy of Sciences}, 113\penalty0 (15):\penalty0 3932–3937, March 2016.
\newblock ISSN 1091-6490.
\newblock \doi{10.1073/pnas.1517384113}.
\newblock URL \url{http://dx.doi.org/10.1073/pnas.1517384113}.

\bibitem[Champion et~al.(2019)Champion, Lusch, Kutz, and Brunton]{Champion_2019}
Kathleen Champion, Bethany Lusch, J.~Nathan Kutz, and Steven~L. Brunton.
\newblock Data-driven discovery of coordinates and governing equations.
\newblock \emph{Proceedings of the National Academy of Sciences}, 116\penalty0 (45):\penalty0 22445–22451, October 2019.
\newblock ISSN 1091-6490.
\newblock \doi{10.1073/pnas.1906995116}.
\newblock URL \url{http://dx.doi.org/10.1073/pnas.1906995116}.

\bibitem[Faroughi et~al.(2025)Faroughi, Mostajeran, Mashhadzadeh, and Faroughi]{faroughi2025scientificmachinelearningkolmogorovarnold}
Salah~A. Faroughi, Farinaz Mostajeran, Amin~Hamed Mashhadzadeh, and Shirko Faroughi.
\newblock Scientific machine learning with {Kolmogorov-Arnold Networks}, 2025.
\newblock URL \url{https://arxiv.org/abs/2507.22959}.

\bibitem[Fiorini et~al.(2025)Fiorini, Flint, Fostier, Franck, Hashemi, Michel-Dansac, and Tenachi]{Fiorini_2025}
Camilla Fiorini, Clément Flint, Louis Fostier, Emmanuel Franck, Reyhaneh Hashemi, Victor Michel-Dansac, and Wassim Tenachi.
\newblock Generalizing the {SINDy} approach with nested neural networks.
\newblock \emph{ESAIM: Proceedings and Surveys}, 81:\penalty0 168–192, 2025.
\newblock ISSN 2267-3059.
\newblock \doi{10.1051/proc/202581168}.
\newblock URL \url{http://dx.doi.org/10.1051/proc/202581168}.

\bibitem[Forootani et~al.(2025)Forootani, Goyal, and Benner]{forootani2023robustsindyapproachcombining}
Ali Forootani, Pawan Goyal, and Peter Benner.
\newblock A robust sparse identification of nonlinear dynamics approach by combining neural networks and an integral form.
\newblock \emph{Engineering Applications of Artificial Intelligence}, 149:\penalty0 110360, 2025.
\newblock ISSN 0952-1976.
\newblock \doi{https://doi.org/10.1016/j.engappai.2025.110360}.
\newblock URL \url{https://www.sciencedirect.com/science/article/pii/S0952197625003604}.

\bibitem[Kaheman et~al.(2020)Kaheman, Kutz, and Brunton]{Kaheman_2020}
Kadierdan Kaheman, J.~Nathan Kutz, and Steven~L. Brunton.
\newblock {SINDy-PI}: a robust algorithm for parallel implicit sparse identification of nonlinear dynamics.
\newblock \emph{Proceedings of the Royal Society A: Mathematical, Physical and Engineering Sciences}, 476\penalty0 (2242), October 2020.
\newblock ISSN 1471-2946.
\newblock \doi{10.1098/rspa.2020.0279}.
\newblock URL \url{http://dx.doi.org/10.1098/rspa.2020.0279}.

\bibitem[Kolda and Bader(2009)]{tensor_decomp}
Tamara Kolda and Brett Bader.
\newblock Tensor decompositions and applications.
\newblock \emph{SIAM Review}, 51:\penalty0 455--500, 08 2009.
\newblock \doi{10.1137/07070111X}.

\bibitem[Lee et~al.(2022)Lee, Trask, and Stinis]{lee2021structurepreservingsparseidentificationnonlinear}
Kookjin Lee, Nathaniel Trask, and Panos Stinis.
\newblock Structure-preserving sparse identification of nonlinear dynamics for data-driven modeling.
\newblock In Bin Dong, Qianxiao Li, Lei Wang, and Zhi-Qin~John Xu, editors, \emph{Proceedings of Mathematical and Scientific Machine Learning}, volume 190 of \emph{Proceedings of Machine Learning Research}, pages 65--80. PMLR, 15--17 Aug 2022.
\newblock URL \url{https://proceedings.mlr.press/v190/lee22a.html}.

\bibitem[Rudy et~al.(2016)Rudy, Brunton, Proctor, and Kutz]{rudy2016datadrivendiscoverypartialdifferential}
Samuel~H. Rudy, Steven~L. Brunton, Joshua~L. Proctor, and J.~Nathan Kutz.
\newblock Data-driven discovery of partial differential equations, 2016.
\newblock URL \url{https://arxiv.org/abs/1609.06401}.

\bibitem[Singh and Mukherjee(2024)]{singh2024learnlearnableadaptiverepresentations}
Arunabh Singh and Joyjit Mukherjee.
\newblock Learn: Learnable and adaptive representations for nonlinear dynamics in system identification, 2024.
\newblock URL \url{https://arxiv.org/abs/2412.12036}.

\bibitem[Steiger and Brachtendorf(2025)]{STEIGER2025427}
Martin Steiger and Hans-Georg Brachtendorf.
\newblock System identification with {SINDy} neural networks for transistor modeling.
\newblock \emph{IFAC-PapersOnLine}, 59\penalty0 (1):\penalty0 427--432, 2025.
\newblock ISSN 2405-8963.
\newblock \doi{https://doi.org/10.1016/j.ifacol.2025.03.073}.
\newblock URL \url{https://www.sciencedirect.com/science/article/pii/S2405896325002903}.
\newblock 11th Vienna International Conference on Mathematical Modelling MATHMOD 2025.

\bibitem[Yonezawa et~al.(2026)Yonezawa, Yonezawa, Yahagi, Kajiwara, Kijimoto, Taniuchi, and Murakami]{Yonezawa_2026}
Ansei Yonezawa, Heisei Yonezawa, Shuichi Yahagi, Itsuro Kajiwara, Shinya Kijimoto, Hikaru Taniuchi, and Kentaro Murakami.
\newblock Sparse identification of nonlinear dynamics with library optimization mechanism: Recursive long-term prediction perspective.
\newblock \emph{IEEE Transactions on Cybernetics}, page 1–14, 2026.
\newblock ISSN 2168-2275.
\newblock \doi{10.1109/tcyb.2026.3652850}.
\newblock URL \url{http://dx.doi.org/10.1109/TCYB.2026.3652850}.

\end{thebibliography}

\end{document}